%% file: main.tex
\documentclass[10pt,twocolumn,letterpaper]{article}

\usepackage[pagenumbers]{cvpr} 


\definecolor{cvprblue}{rgb}{0.21,0.49,0.74}
\usepackage[pagebackref,breaklinks,colorlinks,allcolors=blue]{hyperref}

\usepackage[vlined,linesnumbered,ruled]{algorithm2e}
\usepackage{multirow}
\usepackage{array}
\newcolumntype{C}[1]{>{\centering\arraybackslash}p{#1}}
\def\paperID{*****} 
\def\confName{CVPR}
\def\confYear{2025}

\title{SaSaSaSa2VA: 2nd Place of the 5th PVUW MeViS-Text Track}

\author{
    Dengxian Gong$^{1*}$\quad
    Quanzhu Niu$^{1*}$\quad
    Shihao Chen$^{1*}$\quad
    Yuanzheng Wu$^{1*}$\\
    ~Yikang Zhou$^{1}$\quad
    Tao Zhang$^{1}$\quad
    Haobo Yuan$^{2}$\quad
    Lu Qi$^{1}$\quad
    Shunping Ji$^{1\dag}$ \vspace{3mm}\\
    {$^{1}$Wuhan University\quad
    $^{2}$University of California, Merced}\\
}

\begin{document}
\maketitle
\input{sec/0_abstract}    
\input{sec/1_intro}
\input{sec/2_method}
\input{sec/3_exp}

{
    \small
    \bibliographystyle{ieeenat_fullname}
    \bibliography{main}
}


\end{document}

%% file: sec/0_abstract.tex
\begin{abstract}
\renewcommand{\thefootnote}{}\footnote{$^{*}$Equal contribution.}\footnote{$^{\dag}$Corresponding author.}Referring video object segmentation (RVOS) commonly grounds targets in videos based on static textual cues. MeViS benchmark extends this by incorporating motion-centric expressions (referring \& reasoning motion expressions) and introducing no-target queries.
Extending SaSaSa2VA—where increased input frames and \texttt{[SEG]} tokens already strengthen the Sa2VA backbone—we adopt a simple yet effective target existence–aware verification mechanism, leading to \textbf{S}till \textbf{A}wesome \textbf{SaSaSa2VA} (\textbf{SaSaSaSa2VA}). Despite its simplicity, the method achieves a final score of \textbf{89.19} in the 5th PVUW Challenge (MeViS-Text Track), securing \textbf{2nd} place. Both quantitative results and ablations suggest that this existence–aware verification strategy is sufficient to unlock strong performance on motion-centric referring tasks.
\end{abstract}

%% file: sec/1_intro.tex
\section{Introduction}
\label{sec:intro}
Referring Video Object Segmentation (RVOS) focuses on accurately localizing and segmenting target objects at the pixel level throughout a video, guided by natural language descriptions. With the advancement of multi-modal large language models (MLLMs)~\cite{Qwen-VL, Qwen2-VL, Qwen2.5-VL, bai2025qwen3, qwen35blog, chen2024internvl, chen2024far, chen2024expanding, zhou2025they, internvl3, internvl3_5}, research in this area has shifted from static feature matching to deeper semantic understanding of complex visual and linguistic interactions.

\begin{table}[t!]
    \centering
    \resizebox{\linewidth}{!}{
        \begin{tabular}{r|l|ccc|c}
        \toprule
        Rank & Team & $\mathcal{J\&F}$ & $\mathrm{N\text{-}acc}$. & $\mathrm{T\text{-}acc.}$ & Final \\
        \midrule
        \# 1 & HITsz\_Dragon & 78.97 & 96.15 & 97.59 & 90.91 \\
        \rowcolor{lightgray} {\# 2} & {SaSaSaSa2VA} & {71.06} & {100.00} & {96.52} & {89.19} \\
        \# 3 & tobedone & 71.30 & 96.15 & 98.93 & 88.79 \\
        \# 4 & yahooo & 70.38 & 96.15 & 98.40 & 88.31 \\
        \# 5 & junjie\_zheng & 68.37 & 88.46 & 96.79 & 84.54 \\
        \bottomrule
        \end{tabular}
    }
    \caption{\textbf{Leaderboard of the 5th PVUW Challenge MeViS-Text Track at CVPR 2026.} Our \textbf{SaSaSaSa2VA} team achieves a final score of \textbf{89.19} and wins 2nd place.}
    \label{tab:leaderboard}
\end{table}

\begin{figure*}[t]
  \centering
  \includegraphics[width=0.97\linewidth]{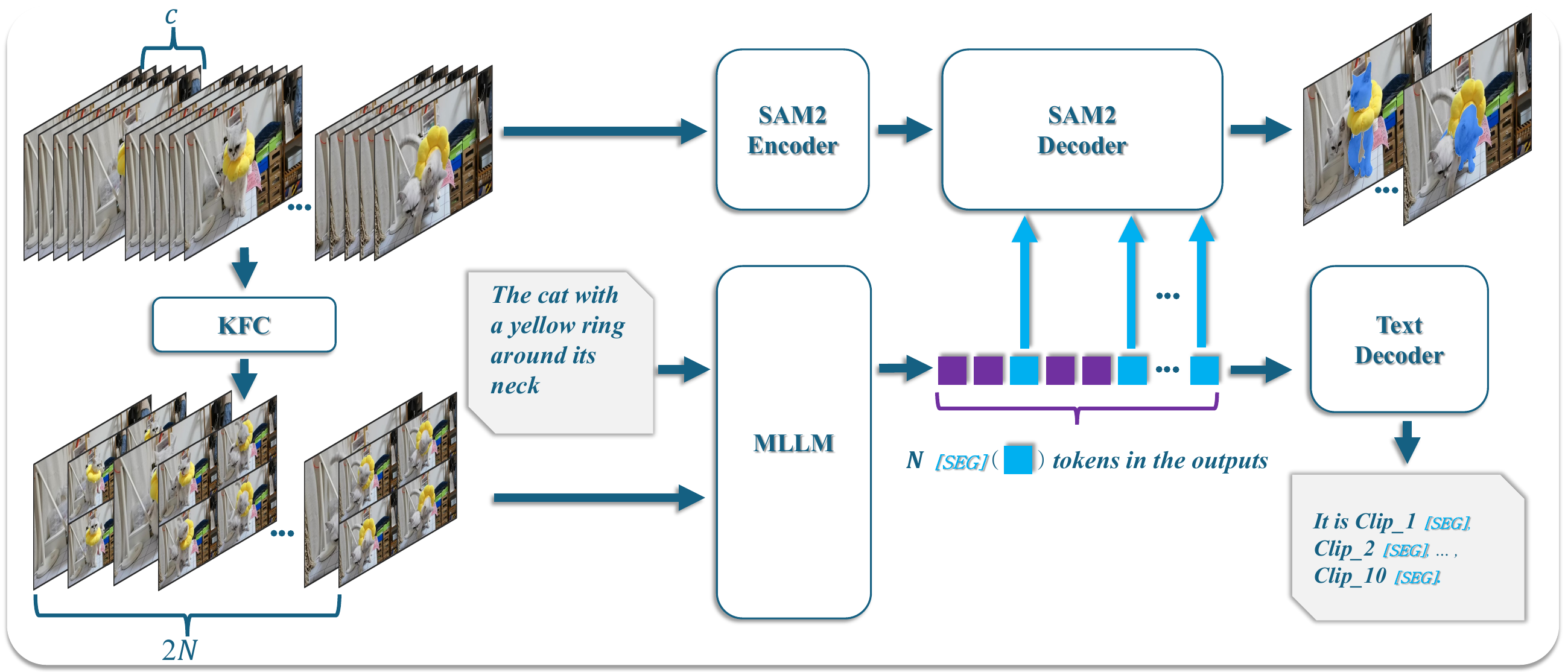}
  \caption{\textbf{The architecture of SaSaSa2VA~\cite{niu20251st}.} 
  Given a video of $T$ frames, the sequence is first split into $N$ temporally ordered clips, each consisting of $c = g^2{+}1$ frames. To improve efficiency while preserving temporal context, frames within each clip are compacted via the Key Frame Compression (KFC) strategy before being fed into the MLLM. Based on the compressed visual inputs, the MLLM generates a set of $N$~\texttt{[SEG]} tokens, where each token encodes segmentation cues for a specific temporal segment. For each clip, SAM2 takes the corresponding \texttt{[SEG]} token as a prompt, together with the original (uncompressed) frames, to decode object masks at the frame level. In this illustration, we set $c=5$, corresponding to $g=2$.}
  \label{fig:sa}
\end{figure*}

The 5th Pixel-level Video Understanding in the Wild (PVUW) Challenge at CVPR 2026 features three highly challenging tracks: Track 1 (MOSEv2), which addresses video object segmentation in complex environments on the MOSEv2~\cite{MOSEv2} dataset; Track 2 and 3, namely MeViS-Text and MeViS-Audio Track, are both based on an enhanced version of the MeViS~\cite{MeViS,ding2025mevis} dataset, focusing on motion-aware referring segmentation from textual descriptions and audio-driven motion segmentation, respectively. Accordingly, this report centers on Track 2 (MeViS-Text Track).

As a benchmark in this domain, the MeViS~\cite{MeViS} dataset has undergone a substantial leap in its latest release, MeViS v2~\cite{ding2025mevis}. Compared to its predecessor, MeViS v2 not only expands the scale but also fundamentally redefines the task landscape.
First, it introduces more challenging motion reasoning expressions, which often involve implicit queries and require models to perform non-trivial logical reasoning. More importantly, MeViS v2 incorporates a large number of no-target expressions, which are particularly deceptive: although semantically aligned with the scene, they do not correspond to any actual object instance. Collectively, these changes necessitate a paradigm shift from conventional conditional localization to a holistic pipeline of target existence verification, reasoning, followed by segmentation and tracking.

In recent years, multi-modal large language models (MLLMs)~\cite{Qwen-VL, Qwen2-VL, Qwen2.5-VL, bai2025qwen3, qwen35blog, chen2024internvl, chen2024far, chen2024expanding, zhou2025they, internvl3, internvl3_5} have rapidly advanced the frontier of visual understanding, enabling comprehensive scene interpretation, fine-grained recognition of objects and actions, and reasoning about interactions among entities across images and videos. Meanwhile, segmentation foundation models have evolved quickly. SAM2~\cite{ravi2024sam} already delivers substantial improvements over prior approaches~\cite{zhang2023dvis, zhang2025dvis++, zhou2024dvis, li2023tube, xu2024rap, zhang20231st_pvuw, zhang20231st_lsvos, niu2025, niu20251st}, largely due to its scalable data engine, which enhances both accuracy and generalization. 
Building on these developments, grounded MLLMs~\cite{lai2024lisa, zhang2024omg, yan2024visa} have shown that instruction-driven segmentation can be effectively realized by integrating MLLMs with specialized segmentation models~\cite{kirillov2023segment, li2024omg, yuan2024mamba}.
More recently, SAM3~\cite{carion2025sam}  further pushes the boundary with a stronger segmentation backbone and introduces an agentic referring segmentation paradigm, enabling more flexible and interactive segmentation processes.

Along this line, Sa2VA~\cite{yuan2025sa2va} combines the advanced MLLM~\cite{chen2024expanding, internvl3, bai2025qwen3} with SAM2~\cite{ravi2024sam}, resulting in a unified framework that delivers strong performance across both visual understanding and segmentation tasks. Based on Sa2VA, the winner of the 7th LSVOS Challenge—SaSaSa2VA~\cite{niu20251st}—addresses the challenges of temporal modeling in long video sequences through Key Frame Compression (KFC) and a multi-\texttt{[SEG]} token strategy, significantly boosting segmentation performance. However, our analysis reveals a critical limitation: despite its excellent performance on standard referring tasks, SaSaSa2VA exhibits a tendency toward forced localization. When encountering the no-target samples in MeViS v2, the absence of an explicit existence verification mechanism often leads the model to produce spurious masklet sequences, resulting in a significant performance bottleneck.

To address this key limitation, we propose \textbf{S}till \textbf{A}wesome \textbf{SaSaSa2VA} (\textbf{SaSaSaSa2VA}). Despite the availability of more powerful MLLM backbones~\cite{bai2025qwen3, qwen35blog, internvl3, internvl3_5}, we leave the SaSaSa2VA architecture intact and avoid costly large-scale retraining. Instead, we perform lightweight fine-tuning on the MeViS v2 dataset to effectively transfer its strong segmentation and temporal modeling capabilities, adapting them to the more challenging motion reasoning requirements introduced in MeViS v2.

Inspired by the “video–language verifier” introduced by the runner-up solution in the previous challenge~\cite{hong2025enhancing, liu2025lsvos2025challengereport}, we adopt a VLM-based inference-time filtering strategy. Specifically, we leverage state-of-the-art vision–language models (VLMs)~\cite{gemini_pro, gpt54} to perform zero-shot target existence verification, which is then used as a filter to refine the outputs of SaSaSa2VA.

Empirically, this hybrid paradigm—fine-tuned segmentation with reasoning-based filtering—effectively handles the negative samples in MeViS v2. As a result, SaSaSaSa2VA achieves a score of \textbf{89.19}, ranking \textbf{2nd} in the challenge, further demonstrating the robustness and competitiveness of SaSaSa2VA~\cite{niu20251st}.

%% file: sec/2_method.tex
\section{SaSaSaSa2VA}
\label{sec:method}
Referring Video Object Segmentation (RVOS) extends traditional object segmentation into the multimodal domain, necessitating the precise localization of a target instance across a temporal sequence guided solely by linguistic cues. Formally, given a video corpus $\mathcal{V} = \{\mathbf{I}_t\}_{t=1}^{T}$ where each frame $\mathbf{I}_t \in \mathbb{R}^{3 \times H \times W}$ denotes an RGB input, and a corresponding linguistic query $\mathcal{T} = \{w_i\}_{i=1}^{L}$ representing a referring expression of $L$ tokens, the objective is to optimize a cross-modal mapping function $f: (\mathcal{V}, \mathcal{T}) \rightarrow \mathcal{M}$. The output $\mathcal{M} = \{\mathbf{M}_t\}_{t=1}^{T}$ is a sequence of pixel-level binary masks, where $\mathbf{M}_t \in \{0, 1\}^{H \times W}$ identifies the spatial extent of the entity described by $\mathcal{T}$ within frame $\mathbf{I}_t$. Our solution consists of two components: the Baseline (\cref{{sec:method_sa2va}}) and the Existence-aware Augmentation strategy (\cref{sec:method_sa}).

\begin{figure}[t]
  \centering
  \includegraphics[width=0.97\linewidth]{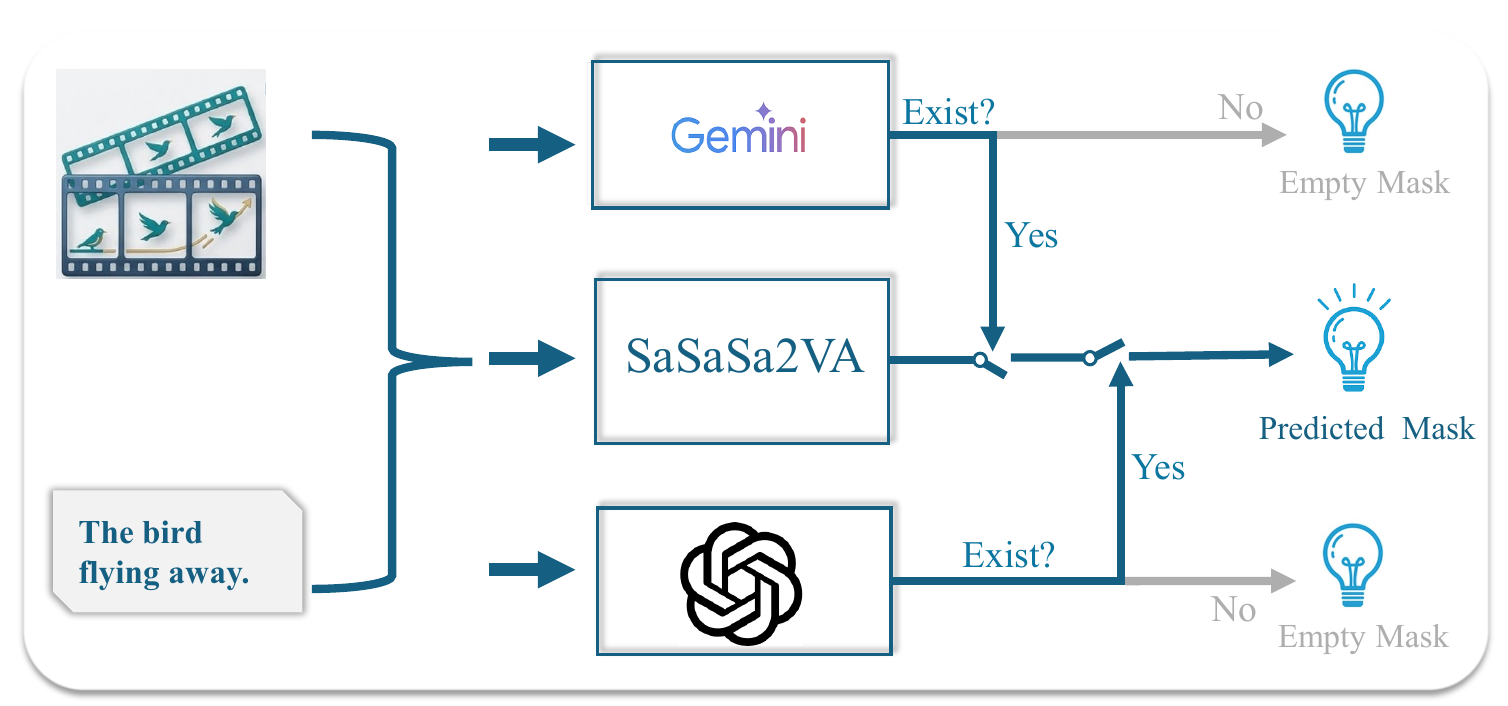}
  \caption{\textbf{The Existence-aware verification illustration of our method.} 
    Given a  video-expression pair $(\mathcal{V}, \mathcal{T})$, \emph{Gemini 3-Flash-Preview} and \emph{GPT-5.4} function as a dual-consensus jury, and an expression is categorized as 'null-target' only under a unanimous consensus, where both models independently confirm the object's absence. Only valid video-text pairs proceed to the SaSaSa2VA base model for inference.}
  \label{fig:ea}
\end{figure}

\subsection{Baseline: SaSaSa2VA}
\label{sec:method_sa2va}

\noindent\textbf{Meta Architecture.}
We adopt SaSaSa2VA~\cite{niu20251st} as our baseline, a unified vision-language segmentation framework that extends Sa2VA~\cite{yuan2025sa2va} with improved temporal modeling and more flexible segmentation interfaces. The model tightly integrates a Multi-modal Large Language Model (MLLM)~\cite{chen2024expanding} with SAM2~\cite{ravi2024sam} as illustrated in~\cref{fig:sa}, enabling end-to-end mapping from multi-modal instructions to pixel-level masks.

Given images, videos, and textual inputs, the MLLM performs cross-modal reasoning and generates structured responses~\cite{lai2024lisa, yan2024visa, zhang2024omg, zhou2025they, zhou2026samtok}. When segmentation is required, the model emits \texttt{[SEG]} tokens, whose hidden representations are treated as implicit prompts for SAM2. This design establishes a direct interface between language reasoning and mask prediction without requiring explicit prompt engineering.

\noindent\textbf{MLLM.}
SaSaSa2VA adopts InternVL 2.5~\cite{chen2024expanding}, following a LLaVA-style architecture~\cite{liu2023visual} composed of an InternViT encoder~\cite{chen2024internvl}, an MLP projector, and a Large Language Model (LLM)~\cite{cai2024internlm2technicalreport, qwen2025qwen25technicalreport}. Visual inputs are encoded into tokens, projected into the language space, and concatenated with text tokens for autoregressive decoding.

Compared to Sa2VA, SaSaSa2VA enhances the interaction between temporal perception and language reasoning via \emph{Segmentation Augmentation}. Instead of relying on sparse frame sampling and a single \texttt{[SEG]} token, the model introduces a more expressive representation that better captures long-range temporal dynamics, without incurring a significant increase in computational overhead. Specifically, it (i) compresses local temporal information into compact representations to increase temporal coverage, and (ii) scales the number of \texttt{[SEG]} tokens to model clip-level variations. As a result, the MLLM can produce multiple segmentation-aware tokens, each corresponding to different temporal segments, improving robustness to object motion, deformation, and occlusion.

\noindent\textbf{SAM2.}
Given the prompt embeddings derived from \texttt{[SEG]} tokens, SAM2~\cite{ravi2024sam} generates high-resolution segmentation masks. Each token serves as a prompt for decoding object masks, which are then temporally propagated to obtain full-video predictions. The use of multiple prompts further enables finer-grained temporal control compared to single-token designs.

To further improve robustness during inference, SaSaSa2VA incorporates multiple sampling strategies at test time (e.g., uniform sampling, content-aware selection, and cyclic coverage), and aggregates the resulting predictions.

\subsection{Existence-aware Augmentation}
\label{sec:method_sa}
Video object segmentation (VOS)~\cite{ravi2024sam,carion2025sam,MOSE,MOSEv2} is a fundamental video tracking~\cite{niu2025,zhang2025dvis++,zhou2024dvis,zhang2023dvis,yang2019video,cheng2021mask2former} task. Distinct from semi-supervised VOS that relies on a ground-truth initial mask, RVOS requires the model to autonomously establish a semantic correspondence between the textual tokens $w_i$ and the visual features of $\mathbf{I}_t$. A critical requirement of the cross-modal mapping function $f: (\mathcal{V}, \mathcal{T}) \rightarrow \mathcal{M}$ is the existence-awareness constraint: in scenarios where the referred object is occluded, absent, or yet to appear in a specific frame $\mathbf{I}_t$, the model must strictly yield a null mask $\mathbf{M}_t = \mathbf{0}$. Consequently, a robust RVOS framework must not only excel in spatial-temporal alignment (captured by $\mathcal{J} \& \mathcal{F}$) but also maintain high discriminative specificity in non-target scenarios, thereby preventing the ``forced-mapping'' hallucinations that typically lead to a catastrophic collapse in Notarget accuracy ($\mathrm{N\text{-}acc}$) performance.

\noindent\textbf{Limitations of SaSaSa2VA.}
Due to the severe lack of negative sample constraints during the training phase of SA2VA~\cite{yuan2025sa2va} and its subsequent variants SaSaSa2VA~\cite{niu20251st}, the models exhibit a pronounced positive inductive bias. In language-guided video segmentation tasks, these models tend to adopt a ``forced mapping'' strategy—attempting to generate a high-confidence masklet regardless of whether the target actually exists in the video sequence. This ``semantic hallucination'' directly prevents the model from outputting an empty mask in null-target scenarios, leading to a precipitous drop in the $\mathrm{N\text{-}acc}$ metric (below 6\%) and severely eroding the global $\mathcal{J} \& \mathcal{F}$ scores due to the surge in false positives as demonstrated by \cref{tab:ablation_study1}.

\noindent\textbf{Overview.} As illustrated in~\cref{fig:ea}, we adopt a target existence–aware verification strategy conditioned on video–language inputs to pre-determine the existence of the referred target in the video and then refine the model predictions accordingly.

\noindent\textbf{Existence-aware verification.}
To rectify the inherent positive bias within the SaSaSa2VA base model, we leverage the high-level semantic reasoning capabilities of state-of-the-art closed-source models, namely \emph{Gemini 3-Flash-Preview}~\cite{gemini_pro} and \emph{GPT-5.4}~\cite{gpt54} as the pre-inference safeguard. Specifically, for each video-expression pair $(\mathcal{V}, \mathcal{T})$, these models function as a dual-consensus jury: all frames $\mathbf{I}_t$ along with the referring expression $\mathcal{T}$ are fed into both models to evaluate the target's presence. We categorize an expression as 'null-target' only under a unanimous consensus, where both models independently confirm the object's absence. In such instances, our framework bypasses the standard segmentation pipeline and preemptively returns a null masklet $\mathbf{M} = \mathbf{0}$. This strategic `consensus gating" effectively shields the system from producing forced-mapping hallucinations, thereby substantially rehabilitating the $\mathrm{N\text{-}acc}$ metric and improving the overall reliability of the segmentation output.

\begin{table}[]
\centering
\begin{tabular}{l|l|ccc}
\toprule
\multicolumn{2}{l|}{Method} & $\mathcal{J\&F}$ & $\mathrm{N\text{-}acc}$  & $\mathrm{T\text{-}acc}$\\
\midrule
\multicolumn{2}{l|}{SaSaSa2VA~\cite{niu20251st}} & \multicolumn{1}{c}{68.04} & \multicolumn{1}{c}{5.26} & \multicolumn{1}{c}{99.77} \\ 
\midrule
\multirow{3}{*}{Ours} & ft.    & 68.55  &  5.26    &   99.77     \\
                    & EA       & 72.43  &  97.37   &    100.00       \\
                    & ft. + EA & \textbf{72.84}  &   \textbf{97.37} &  \textbf{100.00}  \\        
\bottomrule
\end{tabular}
\caption{\textbf{Ablation on Existence-aware Augmentation (EA).} We report the $\mathcal{J\&F}$ and $\mathrm{N\text{-}acc}$ scores of the 26B model on the MeViS V2~\cite{ding2025mevis} valid\_u split. ``ft.'' denotes further fine-tuning on the MeViS V2 training split.}
\label{tab:ablation_study1}
\end{table}



\subsection{Test-time Augmentation}
While SaSaSa2VA~\cite{niu20251st} employs a heavy ensemble mechanism across two dimensions—averaging predictions from multiple sampling strategies (e.g., uniform sampling, content-aware, cyclic) and models of varying scales—our solution adopts a significantly more simple inference pipeline. Specifically, we dispense with both the multi-strategy voting and multi-model aggregation, opting exclusively for a single-model approach powered by the \emph{Uniform+} sampling strategy. For video sequences with an original duration shorter than the training constraint $T$, we maintain temporal coverage by assigning dual \texttt{[SEG]} tokens to specific frames near the clip boundaries. The final segmentation is then derived by averaging the masks from these two corresponding tokens. By focusing on this singular, high-efficiency configuration, we substantially reduce the computational cost while maintaining robust mask generation capabilities.


%% file: sec/3_exp.tex
\section{Experiments}

\subsection{Implementation Details}
Our architectural foundation is anchored by SaSaSa2VA-26B \cite{niu20251st}, an evolution of the Sa2VA-26B framework \cite{yuan2025sa2va}, which harnesses the formidable multimodal reasoning power of the InternVL 2.5-26B backbone \cite{chen2024expanding}. To further push the boundaries of spatial-temporal precision, we subject the model to a rigorous finetuning regimen utilizing the Segmentation Augmentation protocols pioneered by SaSaSa2VA. Under a fixed temporal configuration of $T=100$ and $N=10$ (yielding $c=10, g=3$), we synergize a diverse corpus of grounding data. This includes a trifecta of referring image suites (RefCOCO/+/g \cite{refcoco, refcocog}) interleaved with high-fidelity video benchmarks such as MeViS V2 \cite{ding2025mevis}, Ref-YTVOS \cite{seo2020urvos}, ReVOS \cite{yan2024visa}, and Ref-SAV \cite{yuan2025sa2va}, ensuring the model attains a holistic understanding of dynamic object referring.

\subsection{Main Results}\label{sec:main_result}
The final challenge results benchmarked on the Mevis V2 \cite{ding2025mevis} test split are summarized in Table~\ref{tab:leaderboard}. Remarkably, despite our simple strategy, our methodology achieves a formidable final score of \textbf{89.19}. Most notably, our approach attains a $\mathrm{N\text{-}acc}$ score of \textbf{100.0}, a testament to the efficacy of our dual-consensus existence verification. 

\subsection{Ablation Study}
\label{sec:ablation_study}
\textbf{Existence-aware Augmentation.} 
As evidenced in \cref{tab:ablation_study1}, integrating the Existence-aware Augmentation strategy and further fine-tune the SaSaSa2VA base model yields a substantial performance leap. Notably, it secures a transformative gain of over \textbf{92.11} points in $\mathrm{N\text{-}acc}$ compared to the baseline \cite{niu20251st}, while simultaneously boosting the $\mathcal{J\&F}$ metric by \textbf{4.8} points. This improvement underscores the strategy's efficacy in calibrating the model's existence-detection logic. By suppressing the``forced-mapping" hallucinations typically found in language-guided segmentation, this augmentation enables the framework to distinguish between target presence and absence with high fidelity, thereby refining both temporal consistency and boundary precision.

\section{Conclusion}
\label{sec:conclusion}
This report details our participation in the MeViS-Text track (5th PVUW Challenge), where we focused on mitigating the forced-matching bias inherent in grounded MLLMs. By integrating Existence-aware Augmentation, we successfully prevented the model from erroneously assigning masks to absent targets. Our approach remains straightforward and highly effective, bypassing the need for extensive training or complex multi-stage strategies and yielding a competitive score in the 5th PVUW Challenge